\documentclass[10pt,twocolumn,letterpaper]{article}

\usepackage[pagenumbers]{cvpr} 

\usepackage{graphicx}
\usepackage{amsmath}
\usepackage{amssymb}
\usepackage{booktabs}
\usepackage{times}
\usepackage{epsfig}
\usepackage{overpic}

\usepackage{multirow}
\usepackage{booktabs}
\usepackage{bigstrut}
\usepackage[table]{xcolor}

\usepackage[pagebackref,breaklinks,colorlinks]{hyperref}

\usepackage[capitalize]{cleveref}
\crefname{section}{Sec.}{Secs.}
\Crefname{section}{Section}{Sections}
\Crefname{table}{Table}{Tables}
\crefname{table}{Tab.}{Tabs.}

\def\cvprPaperID{10280} 
\def\confName{CVPR}
\def\confYear{2022}

\begin{document}

\title{Self-attention based anchor proposal for skeleton-based action recognition}

\author{
Ruijie Hou$^{1}$,
Ningyu Zhang$^{1}$,
Yulin Zhou$^{1}$,
Xiaosong Yang$^{2}$,
Zhao Wang $^{1*}$\\
$^1$Zhejiang University,
$^2$Bournemouth University\\
{\tt\small zwang@outlook.com}
}

\maketitle

\begin{abstract}
Skeleton sequences are widely used for action recognition task due to its lightweight and compact characteristics. Recent graph convolutional network (GCN) approaches have achieved great success for skeleton-based action recognition since its grateful modeling ability of non-Euclidean data. GCN is able to utilize the short-range joint dependencies while lack to directly model the distant joints relations that are vital to distinguishing various actions. Thus, many GCN approaches try to employ hierarchical mechanism to aggregate wider-range neighborhood information. We propose a novel self-attention based skeleton-anchor proposal (SAP) module to comprehensively model the internal relations of a human body for motion feature learning. The proposed SAP module aims to explore inherent relationship within human body using a triplet representation via encoding high order angle information rather than the fixed pair-wise bone connection used in the existing hierarchical GCN approaches. A Self-attention based anchor selection method is designed in the proposed SAP module for extracting the root point of encoding angular information. By coupling proposed SAP module with popular spatial-temporal graph neural networks, e.g. MSG3D, it achieves new state-of-the-art accuracy on challenging benchmark datasets. Further ablation study have shown the effectiveness of our proposed SAP module, which is able to obviously improve the performance of many popular skeleton-based action recognition methods. Related code will be available on \url{https://github.com/ideal-idea/SAP}
\end{abstract}

\section{Introduction}
\label{sec:intro}

Skeleton based human 3D action recognition aims help machines understand human behaviors which is a core challenge due to the multiple granularities and large variation of given motions based on human body-skeleton. It is a fundamental task in human-centred scene understanding which has attracted increasing attention to the community. With the development of low-cost motion sensors and effective human pose estimation techniques, skeleton-based motion representation become more compact and robust to environment compare with its video counterpart. In addition, skeleton-based motion representation conveys relatively high-level information that exhibits robustness to appearance variation such as background clutter, illumination changes. The related action recognition techniques are widely applied to many applications \cite{2010/SurveyVisionbasedHuman}, such as computer such as video surveillance, human-computer interaction, video retrieval and autonomous driving.

Skeleton based human motion data is a time series of joints' 3D coordinates that either estimated from 2D videos with pose estimation methods \cite{cao2017realtime,cao2019openpose} or directly collected by sensors, e.g. Kinect, MoCap System and wearable IMUs. In oder to extract discriminate features for skeleton-based action recognition, great effort has been made to learn patterns from spatial configuration and temporal dynamics of joints. Early approaches could be mainly categorized into Recurrent Neural Network (RNN) based or Convolutional Neural Network (CNN) based. The skeleton based action data is manually modified to a kind of grid-shape structure\cite{kim2017interpretable} or a sequence of the coordinate vectors\cite{du2015hierarchical,zhang2017view}. For instance, a hierarchical bidirectional RNN has been employed to capture dependencies within body parts \cite{du2015hierarchical}. A trimmed skeleton sequences have been used in a CNN architecture for action classification \cite{du2015skeleton,li2017skeleton}. However, aforementioned methods fail to fully explore the inherent relationships between joints due to natural graph structure of skeleton. In order to address such drawbacks, Yan el al. have adopted a graph convolutional network (GCN) based method named ST-GCN \cite{yan2018spatial} for modeling dynamic skeleton sequences. A spatial temporal skeleton graph is constructed with the joints as graph as nodes, natural connections of body and time as edges. Thus, an adjacency matrix of the skeleton graph is built that contains spatial temporal relations between joints.

\begin{figure}[t!]
\center
   \begin{overpic}[width=0.8\columnwidth]{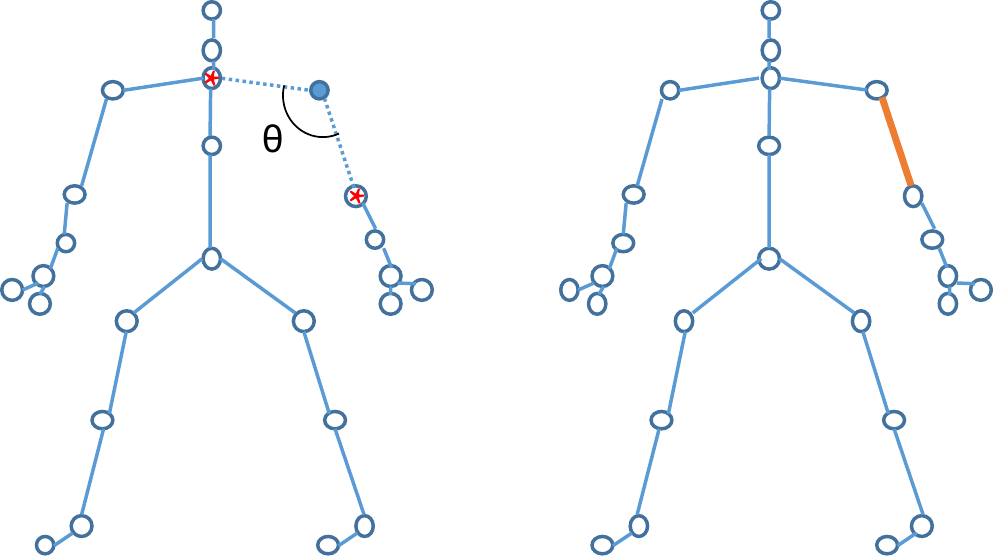}
    \end{overpic}
    \caption{Example of triplet angle representation and bone connection: (Left) Angle representation, where the blue point is the target joint and red stars are anchors;(Right) Bone connection, where orange line is the connection of selected pair of joints.
    }\label{fig:angle-bone}
\end{figure}

The follow up works have presented further effective GCN based models for skeleton based action recognition \cite{li2019spatio,si2019attention,shi2019skeleton}.  These work has tried to either capture both short-range and distant joints relations in the spatial domain or consider both the short-term trajectory and long-term trajectory that are essential to similar action recognition. Further research has conducted multi-scale ST-GCN based approaches while indicating that using only the coordinates of joints is less efficient to explore the structures of skeletons \cite{xia2021multi,chen2021multi}. How ever, such pair-wised representation of fixed topology constraint, e.g. bone connections, ignores the implicit joint correlations.  On the other hand, existing work has revealed that exploring angular information could improve the classification performance among actions with similar trajectories \cite{yan2018spatial,cheng2020skeleton,cheng2020decoupling,song2021constructing}. An example of angle representation and bone connection is shown in Fig \ref{fig:angle-bone}. Therefore, a skeleton-anchor proposal module (SAP) is designed in this paper to explore the angle information, which is a high order triplet representation of skeleton joint topology. It needs to point out that the distinctive features are usually contained in a limited temporal scale and body parts, especially for similar action distinguish. Hence, a self-attention based method is designed in the proposed SAP to automatically extract the skeleton-anchors, the root points for encoding target joint's angle information, which aims to enhance the contextual information extraction from informative body parts and frames.  In addition, we use receptive weight to adjust the range of anchor location, where the the anchors are estimated in learned location around body rather than the manually specified strategy used in \cite{qin2021fusing}.

To verify the superiority of the proposal SAP module, extensive experiments are performed on the challenging benchmark dataset NTU-RGB+D \cite{shahroudy2016ntu}. By coupling with popular spatial-temporal graph neural networks, e.g. MSG3D, it outperforms state-of-the-art works. The ablation study of proposed SAP module show the effectiveness of anchor selection strategy. The main contributions of this work are summarized as follow:

\begin{itemize}
  \item A high order triplet representation that encoding angle information is designed in this paper to extract discriminate semantic spatial-temporal features rather than the traditional pair-wise bone connection, which is able to extract complementary feature to current ST-GCN method.
  \item A self attention based anchor location learning method is designed in the SAP module to automatically determine the anchor location with given motion, where the anchor position could be located around body part. The ablation study has shown the advantage of proposed method compare with the manually specified strategy.
  \item We incorporate the proposed SAP module into the existing ST-GCN module, e.g. MS-G3D, and achieve state-of-the-art results on benchmark dataset. Meanwhile, the proposed SAP module can be easily adopted to improve the performance of existing approaches since it is able to provide complementary information to current joint and bone representation.
\end{itemize}

\section{Related Work}
\label{sec:rw}
Action recognition based on skeleton data has received lots of attention due to its robustness against variation of background and compactness to the RGB based represent. Handcrafted features were used in early approaches, where features could be manually designed based on joint angles \cite{ofli2014sequence}, kinematic features\cite{zanfir2013moving}, trajectories \cite{wang2016graph} or their combinations \cite{wang2016adaptive}. With the development of deep learning, Many CNN or RNN based data-driven methods that could automatically learn the action patterns have attracted much attention. For instance, the skeleton action could either be treated as motion images in CNN approaches \cite{ke2017new,kim2017interpretable} or modeled as sequences of coordinates in RNN approaches\cite{du2015hierarchical,liu2016spatio}.

Graph-based methods have drawn much attention in recent skeleton based action recognition methods due to its ability of modeling relationship between body joints. The first GCN model for skeleton based action recognition is ST-GCN proposed by \cite{yan2018spatial}, where the skeleton is treated as a graph, with joints as nodes and bones as edges. Some data dependent methods have been developed to further capture the relationship between distant joints \cite{shi2019skeleton,li2019spatio,zhang2020context}. In addition, the multi-scale structural feature representation method have been developed  via higher order polynomials of the skeleton adjacency matrix. For instance, a multiple-hop module is used to break the limitation of representational capacity ca caused by first-order approximation \cite{peng2020learning,liu2020disentangling,xia2021multi}. Inspired by \cite{liu2020disentangling}, a sub-graph convolution cascaded by residual connection with enrich temporal receptive field is proposed by \cite{chen2021multi}. A combination approaches called Efficient GCN is designed by \cite{song2021constructing}. An multi-granular GCN based methods on temporal domain is designed by \cite{chen2021learning}. Angle information extracted from manually specified joint groups is fused to GCN model in \cite{qin2021fusing}.
\begin{figure*}[t!]
    \center
   \begin{overpic}[width=0.8\linewidth]{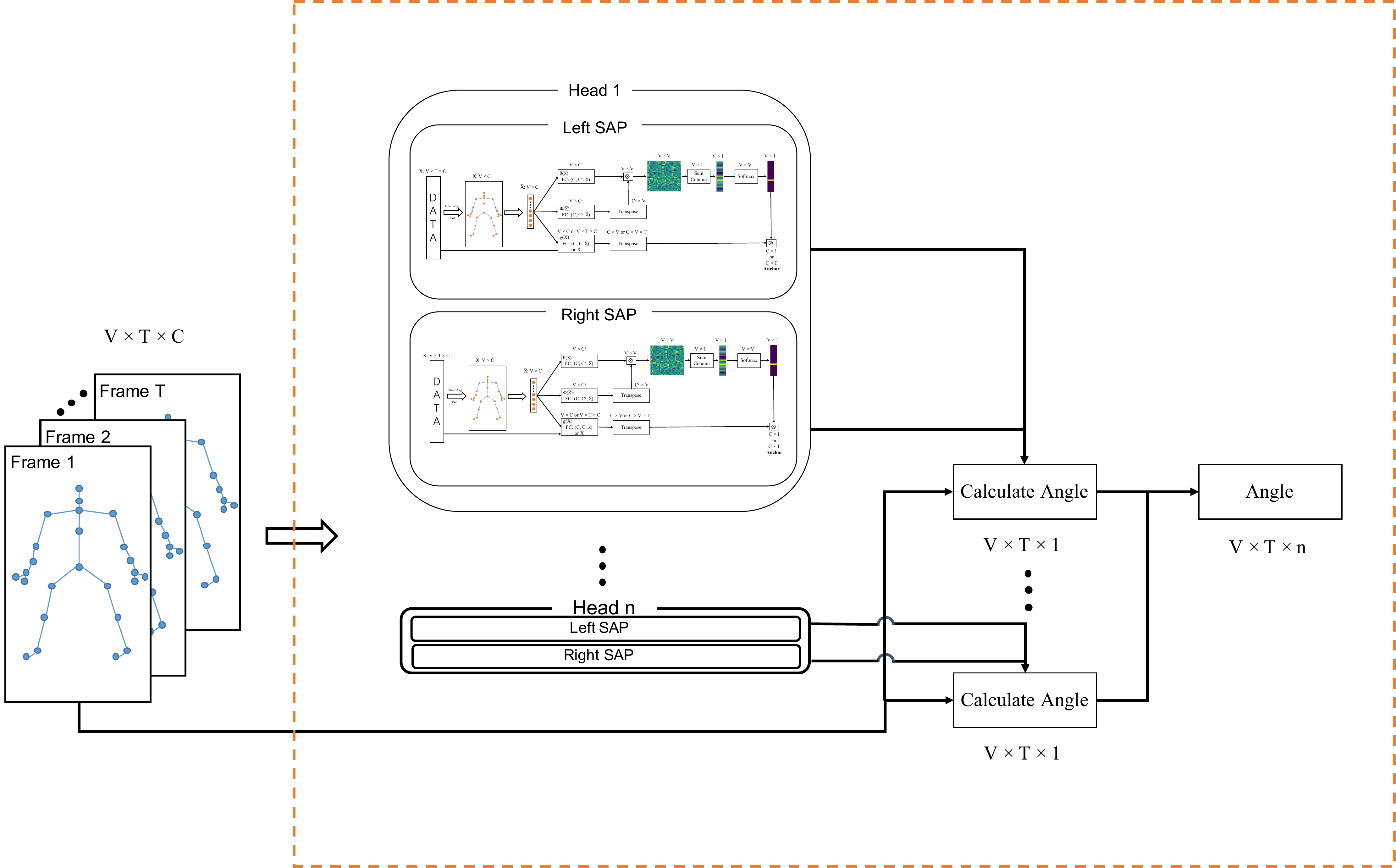}
    \end{overpic}
    \caption{
    Use the multi\_head mechanism to obtain multiple angle information. The orange box is our gen\_angle module, which uses the Eq \eqref{equ:gen_angle} to calculate the angle.
    }\label{fig:generate_angle_multi_head}
\end{figure*}
\section{Methodology}\label{sec:meth}
\subsection{Angle relation representation}
\begin{figure}[t!]
\center
   \begin{overpic}[width=0.6\columnwidth]{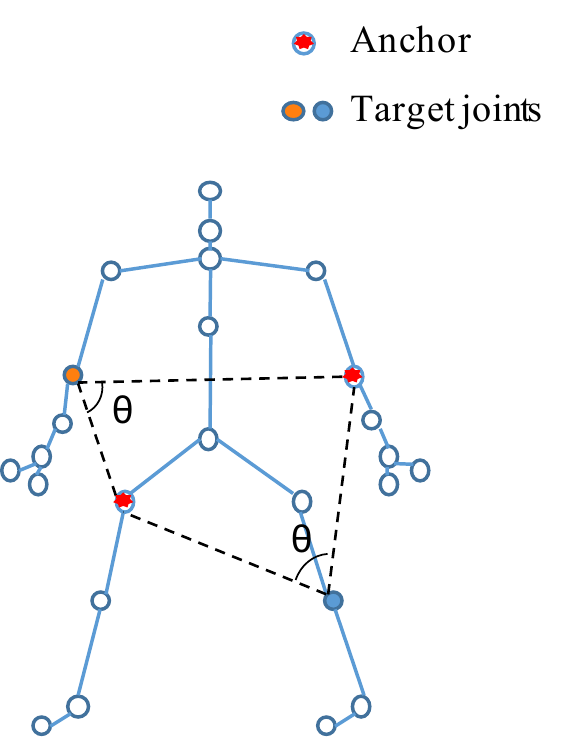}
    \end{overpic}
    \caption{Obtain the angle information according to the anchor. The orange and blue node are target points $w_1$, $w_2$, and the red star nodes are the anchors that is used to calculate the angle.
    }\label{fig:generate_angle}
\end{figure}

In order to extract triplet representation via encoding angle information, we calculate the angles between other nodes and these two anchors. Assume a pair of anchors are denoted as $w_1$,$w_2$ and joint node is denoted u,  $\vec{b}_{u w_{i}}$represents the vector from node $u$ to node $w_i$ ($i=1,2$). Let $\left(x_{k}, y_{k}, z_{k}\right)$ represents the coordinate value of the $k$th point, then $\vec{b}_{u w_{i}}$ could be calculated as $\vec{b}_{u w_{i}}=\left(x_{w_{i}}-x_{u}, y_{w_{i}}-y_{u}, z_{w_{i}}-z_{u}\right)$.

As shown in Fig \ref{fig:generate_angle}, the angle of the node $u$ could be obtained through  Eq \eqref{equ:gen_angle}. In this way, the angle information of each node with given pair of anchors can be obtained. Note that $w_1$ and $w_2$ are not required to be adjacent nodes of node $u$. Compare with bone connection that focus on the length constraint and directions, such angle information focus more on fine grained relative movement and is invariant to the body scale.
\begin{equation}\label{equ:gen_angle}
    d_{a}(u)=\left\{\begin{array}{ll}
\cos \theta=\frac{\vec{b}_{u w_{1}} \cdot \vec{b}_{u w_{2}}}{\left|\vec{b}_{u w_{1}}\right|\left|\vec{b}_{u w_{2}}\right|} & \text { if } u \neq w_{1}, u \neq w_{2}, \\
0  & \text {if } u=w_{1} \text { or } u=w_{2} .
\end{array}\right.
\end{equation}
\begin{figure*}[htbp]
\center
   \begin{overpic}[width=0.95\linewidth]{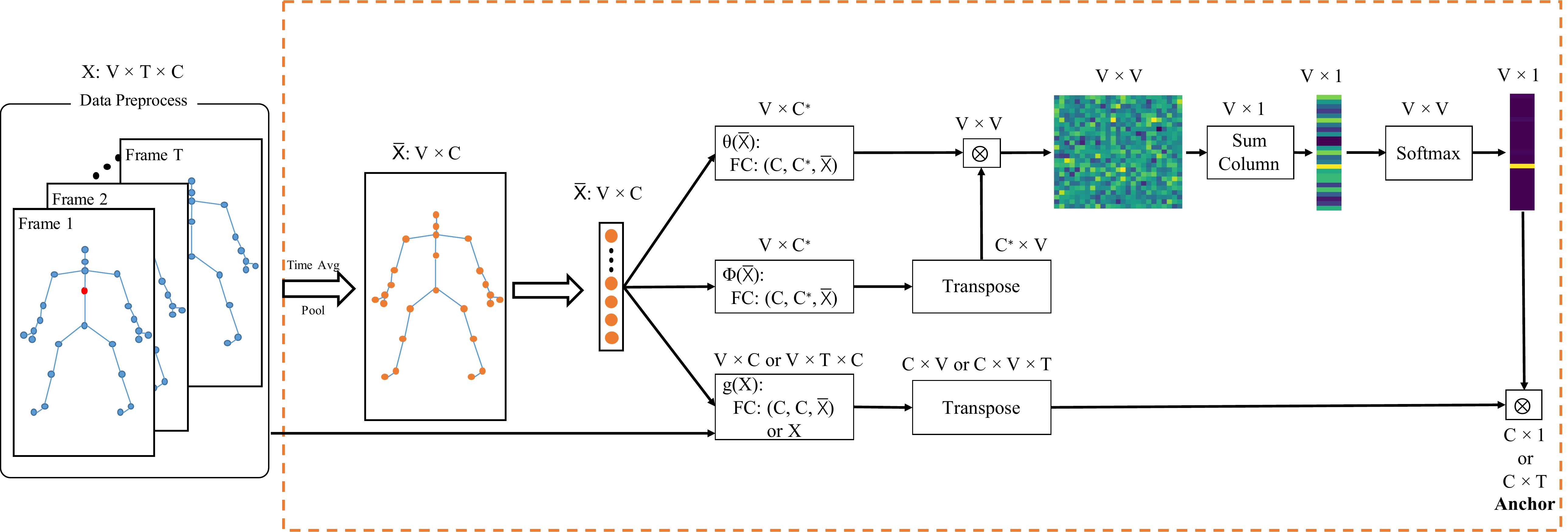}
    \end{overpic}
    \caption{Using the attention mechanism to select anchor points, the orange box is the SAP module
    }\label{fig:sap}
\end{figure*}

In order to capture the dynamic relationship of joints in different granularity of relative movement scale,  multiple pairs of anchors are demanded. The computational cost could be high if all possible combinations of $(u,w_{i})$ are employed. Hence, we need to define a compact sets of anchors to facilitate distinguishing actions while avoiding massive growth of computational cost. Anchors would be generated through the SAP module in this work. The overall procedure is shown in Fig \ref{fig:generate_angle_multi_head}. Anchors will selected with dual SAP modules, and then the angle information according to different anchor pairs would be calculated for each node. Besides, it needs to point out that no information would be provided while the pair of anchors are same according to Eq \eqref{equ:gen_angle}. Detail of meaningful anchors generation would be introduce in the following paragraph.

\subsection{Anchor Generation}

The proposed SAP module employs a modified self-attention mechanism to select the joint called anchor., which is shown in Fig \ref{fig:sap}. In order to reduce the complexity of the model, we set the input used to calculate the similarity in the SAP module to be the average of the original input in time dimension. Let $\bar x_i$ denotes the average value of the coordinates of the i-th node in all frames, it is calculated with Eq \eqref{equ:x},
\begin{equation}\label{equ:x}
    \bar x_i = \frac{\sum_{t=1}^{t=T}x_{i,t}}{T}
\end{equation}
where $\bar x_i\in \mathbb{R}^3$.

In the SAP module, it would firstly obtain the sum of the similarity between the $i$th node and each node with Eq \eqref{equ:fx} and Eq \eqref{equ:cx}.

\begin{equation}\label{equ:fx}
    f(\bar x_i,\bar x) = \mathrm{e}^{\sum_{\forall{j}}\theta(\bar x_i)\times\phi(\bar x_j)^T}
\end{equation}
\begin{equation}\label{equ:cx}
    C(\bar x)=\sum_{\forall{i}}f(\bar x_i,\bar x)
\end{equation}

Then \emph{softmax} is applied to the weight summation results of all nodes to generate the normalized weights of each node, $wight\_i$, as Eq \eqref{equ:weight}.
\begin{equation}\label{equ:weight}
weight_i=\frac{1}{C(\bar x)}\times f(\bar x_i, \bar x)
\end{equation}

After that, the position obtained by the weighted sum of all nodes is used as the anchor point, which is defined as Eq \eqref{equ:center}:
\begin{equation}\label{equ:center}
    center_{i,t} = \sum_{\forall i}weight_i \times g(x_{i,t})
\end{equation}

Considering the component $f(x_i)$ in Eq \eqref{equ:fx} that gets the sum of the similarity among the $i$th node and each node, such step has one more summation operation than the original self-attention method.
The reason is that the SAP module aims to choose an anchor point, rather than generating new features for each node.

Instead of simply doing the dot product on the input directly, Eq \eqref{equ:thetax} and Eq \eqref{equ:phix} is used to conduct further transform of the input and then estimate the similarity. The advantage of this is that it can distinguish the center point and the adjacent point of the calculating similarity operation. And it will not be limited by the similarity in the geometric sense of the input.

\begin{equation}\label{equ:thetax}
\theta(\bar x)=w_{\theta}\times \bar x
\end{equation}

\begin{equation}\label{equ:phix}
\phi(\bar x)=w_{\phi}\times \bar x
\end{equation}

\subsection{Anchor location range}

A further attempt of varying the range of anchor location is conducted in this study. Regarding the position range where the anchors can fall, we compared three choices: the range of anchors is limited, the anchor would fall on the original body joints; the range of anchors is relaxed, the anchors could be located inside the original body, and then the range of anchors is further relaxed, the anchors could be located around the original body. This is achieved by the add a control component of $\alpha$ to Eq \eqref{equ:fx} and adjusting $g(x)$ in Eq \eqref{equ:center}, where Eq \eqref{equ:fx} is rewritten as Eq \eqref{equ:fx2}

\begin{equation}\label{equ:fx2}
    f(\bar x_i,\bar x) = \mathrm{e}^{\sum_{\forall{j}}\theta(\bar x_i)\times\phi(\bar x_j)^T\times\alpha}
\end{equation}

The $\alpha$ in Eq \eqref{equ:fx2} is used to control the degree of dispersion of the Softmax generation weight. The larger the $\alpha$, the larger the maximum value of Softmax and the smaller the minimum value, which makes the generated anchor point is more inclined to be located on a origin joint.

The $g(x)$ in Eq \eqref{equ:center} is used to control whether the anchor point falls within the body or around the body. It is can also adjust using same anchors or update anchors for each frame.

\paragraph{The anchor should located on joint or within the body}
The the anchor generated by the SAP module should be in the body or on the joints.
\paragraph{implementation detail}
Let $g(x_{i,t})=x_{i,t}$, the anchor is generated by weighting each joint point, so the anchor point will fall within the body range, and each frame will generate its own anchors. When $g(x_{i,t})=x_{i,0}$, Each frame uses the anchors of the first frame, so that the angle information can take into account some global position information.And when the value of $\alpha$ is large, the maximum value of the weight generated by Softmax is as close to 1 as possible, and the minimum value is as close to 0 as possible. At this time, the weighted sum can achieve the effect of selecting the most salient node among the original nodes, which can make the generated anchors falls on the original joints as much as possible.
\paragraph{The anchor could be located around body}
The anchors generated by self-attention is no longer in the natural body range, and can be selected arbitrarily in the entire coordinate system. The purpose of this is to find a better anchor position than within the body.
\paragraph{implementation detail}
Let $g(x_{i,t})=w_g \times \bar x_i$. Linear transformation of the original coordinates and weighting can make the anchors generated by SAP fall around body. At this time, the same anchor points are used in every frame.
\begin{figure*}[htbp]
    \centering
    \includegraphics[width=1\linewidth]{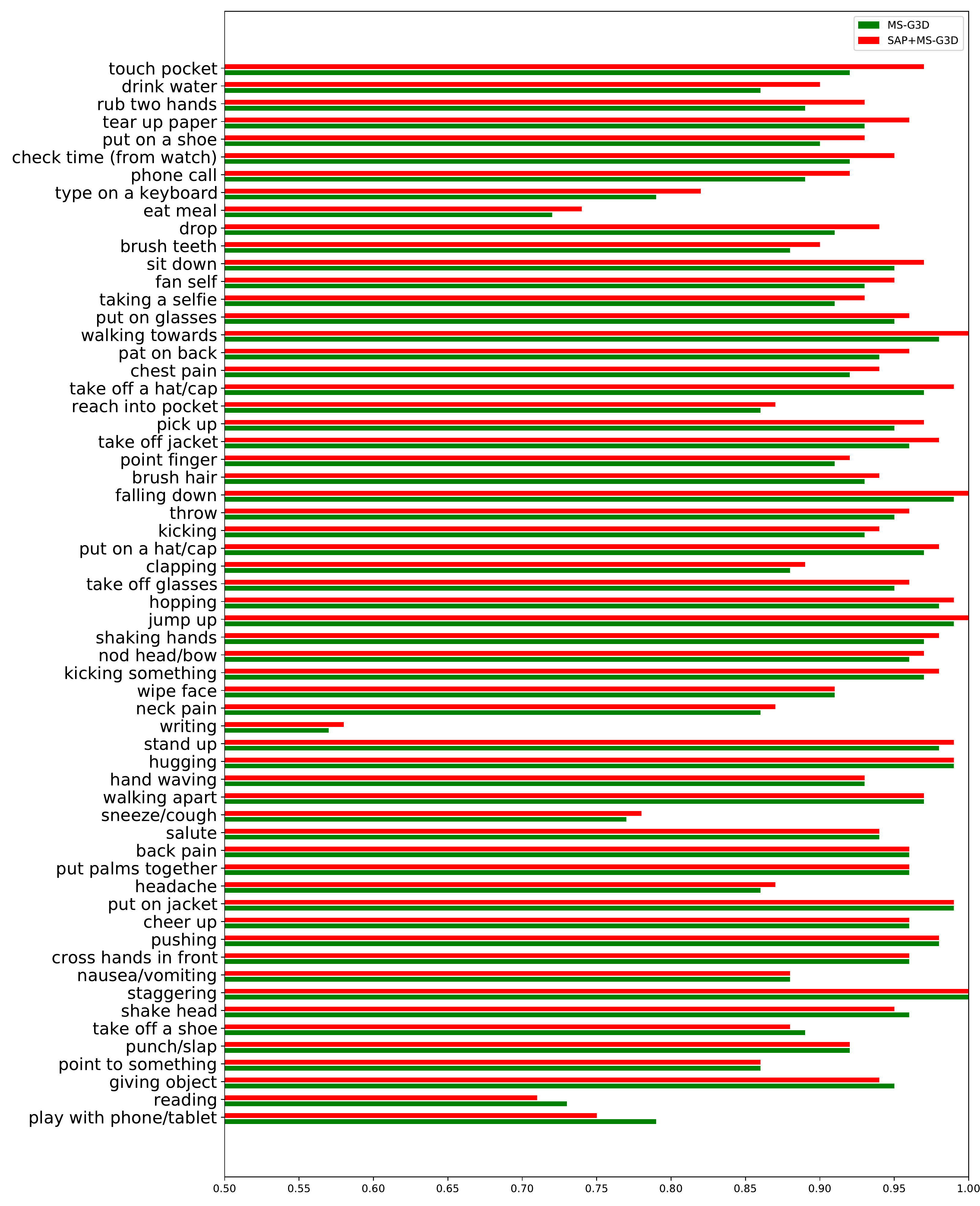}
    \caption{A comparison of action recognition for MS-G3D w/o SAP on each class of NTU RGB+D}
    \label{fig:detail_diff}
\end{figure*}
\begin{figure*}[htb]
    \centering
    \includegraphics[width=1\linewidth]{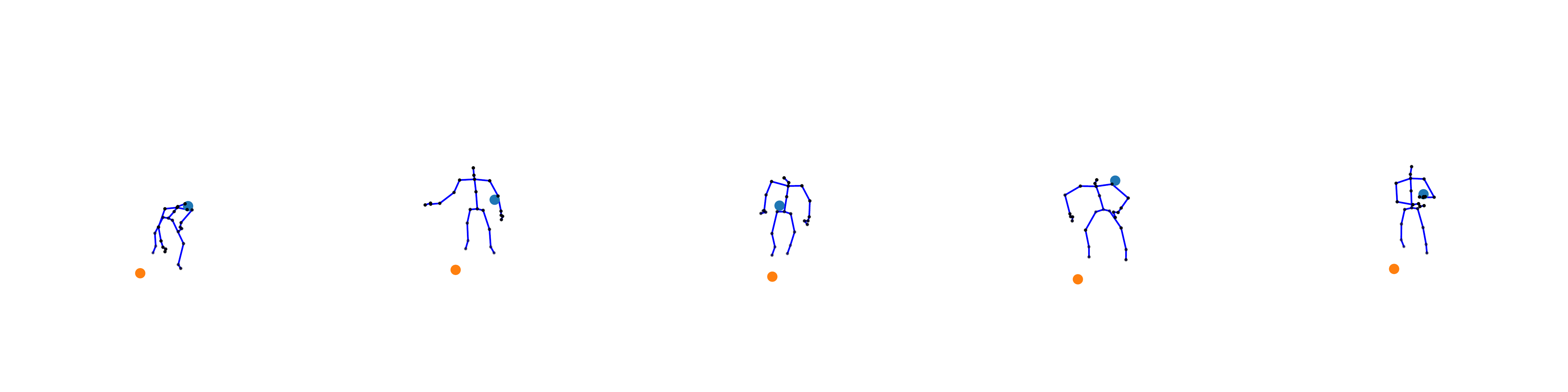}
    \caption{Example of Anchors location around body}
    \label{fig:mylti-g-different-samples}
\end{figure*}

\section{Experimental Results and Discussion}\label{sec:exp}
In this section, we evaluate the proposed SAP module on benchmark dataset NTU RGB+D \cite{shahroudy2016ntu}. Ablation study is taken to validate the contribution of module component. For simplicity, the ablation study is firstly taken with a simple resnet-18 based model to determine hyper-parameter while MS-G3D \cite{liu2020disentangling} is employed as baseline for the performance evulation. We train deep learning models on 2 NVIDIA RTX 3090 GPUs and use PyTorch as our deep learning framework . Furthermore, we apply stochastic gradient descent (SGD) with momentum 0.9 and learning rate 0.05 with step of 10 times decay at the 30th and 40th epoch.

As shown in Table \ref{tab:ablation-anchor-location}, we finally choose the strategy that get new anchor locations in the range of around body.

\subsection{Datasets}

NTU RGB+D: NTU RGB+D \cite{shahroudy2016ntu} dataset contains 60 different human action classes and it is the most widely used dataset for evaluating skeleton-based action recognition models. It consists of 56,880 action samples in total which are performed by 40 distinct subjects. The 3D skeleton data is collected by Microsoft Kinect v2 from three cameras simultaneously with different horizontal angles: -45, 0, 45. The human pose in each frame is represented by 25 joints. We follow the two official evaluation protocols for performance evaluation: Cross-Subject (X-sub) and Cross-View (Xview). Under the former protocols, half of the 40 subjects consists of the training set and the other for testing. For the latter, samples captured by camera 2 and 3 are used for training and the rest are for testing.

\subsection{Ablation Study}
All experiments in this section are conducted on NTU-60 dataset and follow linear evaluation protocol of  Cross-Subject (X-sub).

\paragraph{Effeteness of Number of Anchors} The number of anchors is corresponding to the number of multi-head in the SAP. The result of ablation study on number of anchors is presented in Table \ref{tab:ablation-num-head}. The first row indicate the manually specified 7 anchors with same criteria used in \cite{qin2021fusing}. With the increasing on number of anchors, the performance is also increased. Considering the trade-off between effectiveness and efficiency, we finally determine a 5-head structure for further performance evaluation.
\begin{table}[htbp]
  \centering
  \caption{Ablation Study on Number of Anchors}
  {
    \begin{tabular}{lccc}
    \toprule
    \multicolumn{1}{c}{\multirow{2}[4]{*}{Methods}} &
    \multicolumn{1}{c|}{\multirow{2}[4]{*}{backbone}} &
    \multicolumn{2}{c}{NTU60} \\
\cmidrule{3-4} & \multicolumn{1}{c|}{} & X-Sub & Acc $\uparrow$ \\
    \midrule
    fixed 7 & resnet18     & 77.1  & - \\
    one head & resnet18     & 75.84 & - \\
    three head & resnet18   & 82.38 & 5.28 \\
    \textbf{five head} & resnet18      & 84.4  & 7.3 \\
    seven head & resnet18      & 84.32 & 7.22 \\
    ten head & resnet18      & 84.27 & 7.17 \\
    fifteen head & resnet18      & 84.81 & 7.71 \\
    \bottomrule
    \end{tabular}
  \label{tab:ablation-num-head}
  }
\end{table}

\paragraph{Effectiveness of Anchor Location Constraint} Anchors could be  generated using first frame as reference or updated in each frame. In addition, the hyper parameters $\alpha$ and $g(x)$ are used to control the range of anchor location For instance, the anchors would be limited on joints with a large $\alpha$, e.g. $20$ in our experiment. An example of anchors located aound body is shown in Fig \ref{fig:mylti-g-different-samples}. We noticed that the learning anchors with less limitation could achieve better performance.

\begin{table}[htbp]
  \centering
  \caption{Ablation Study on Anchor Location Constraint}
  {
    \begin{tabular}{lccc}
    \toprule
    \multicolumn{1}{c}{\multirow{2}[4]{*}{Methods}} &
    \multicolumn{1}{c|}{\multirow{2}[4]{*}{backbone}} &
    \multicolumn{2}{c}{NTU60} \\
\cmidrule{3-4}
& \multicolumn{1}{c|}{} & X-Sub & Acc $\uparrow$ \\
    \midrule
    \multicolumn{4}{c}{Extract anchor for each frame} \\\hline
    fixed 7 Joints & resnet18     & 77.1  & - \\
    Anchors on Joints & resnet18     & 72.2  & - \\
    Anchors within Body & resnet18     & 78  & 0.9 \\
    \hline \multicolumn{4}{c}{Using anchor extracted from 1st frame} \\\hline
    fixed 7 Joints & resnet18     & 75.9  & - \\
    Anchors on Joints & resnet18     & 78.5  & 1.4 \\
    Anchors within Body & resnet18     & 78.7  & 1.6 \\
    \hline \multicolumn{4}{c}{Using anchor extracted from global frame space} \\\hline
    Anchors around Body & resnet18     & 84.4  & 7.3 \\
    \bottomrule
    \end{tabular}%
  \label{tab:ablation-anchor-location}%
  }
\end{table}%

\begin{figure*}[htbp]
    \centering
    \includegraphics[width=1\linewidth]{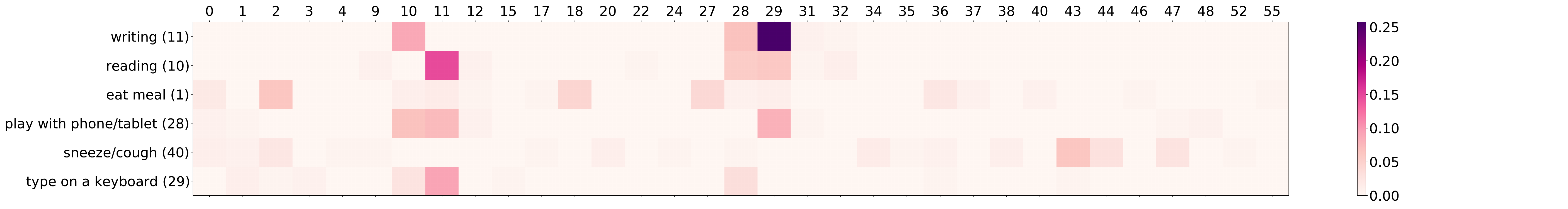}
    \caption{Confusion matrix of our method with failure actions (less than $85\%$ accuracy on X-sub benchmark), where the number in coordinate axes represent the action index, the color stands for the misclassification rate}
    \label{fig:confuse}
\end{figure*}

\subsection{Comparison with State-of-Art}
We compare the model accuracies of our method against previous SOTA methods on both on both X-sub and X-view benchmarks. The results are displayed in Table \ref{tab:sota}, There are several typical comparisons shown as follows:

From Table \ref{tab:sota}, our method's accuracy on X-sub benchmark is $92.5\%$ and X-view benchmark is $96.6\%$, which outperforms all other SOTA method. Compare with the baseline MS-G3D \cite{liu2020disentangling}, our method leads an obvious improvement. Compare with current SOTA, e.g. EfficientGCN\cite{song2021constructing} and MG-GCN\cite{chen2021learning}, the performance of our method is slightly better, where additional velocity/temopral information is used in their works.

These results imply that the proposed SAP module is able to conduct a competitive performance compared to the SOTA. We consider that our proposed SAP module is able to provide complementary information via exploiting the multi-scale relative movement knowledge via encoding angle information. Moreover, the attention mechanism is able to makes the model prone to discover the informative joints with a complementary view from angle information.

\begin{table}[htbp]
  \centering
  \caption{Comparison with State-of-Art}
    {
    \begin{tabular}{lccc}
    \toprule
    \multicolumn{1}{c}{\multirow{2}[4]{*}{Methods}}  & \multicolumn{1}{c|}{\multirow{2}[4]{*}{Publisher}} & \multicolumn{2}{c}{NTU60} \\
\cmidrule{3-4}          &    \multicolumn{1}{c|}{}   & \multicolumn{1}{c|}{X-Sub} &X-View  \\
    \midrule
    ST-GCN\cite{2018/SpatialTemporalGraph}  & AAAI18       & 81.5 & 88.3\\
    2s-AGCN\cite{2019/TwoStreamAdaptiveGraph}& CVPR19         & 88.5 & 95.1\\
    DGNN\cite{shi2019skeleton}  & CVPR19         & 89.9& 96.1 \\
    DSTA-Net\cite{2020/DecoupledSpatialTemporalAttentionb}   & ACCV20     & 91.5 & 96.4 \\
    MS-G3D\cite{liu2020disentangling} &     CVPR20      & 91.5 & 96.2\\
    4s Shift-GCN\cite{cheng2020skeleton} &CVPR20 &90.7 &96.5\\
    MST-GCN\cite{chen2021multi}  &  AAAI21  & 91.5 & 96.6\\
    EfficientGCN\cite{song2021constructing} &  arxiv  & 91.7 & 95.7\\
    AngNet\cite{2021/FusingHigherOrderFeatures}    &  arxiv  & 91.7 & 96.4\\
    MG-GCN\cite{chen2021learning}  &  MM21  & 92.0 & 96.6\\

    \midrule
    \multicolumn{3}{c}{Our Methods} \\
    \midrule
    SAP + MS-G3D    &6   & \textbf{92.5} & \textbf{96.9}\\

    \bottomrule
    \end{tabular}%
    }
  \label{tab:sota}%
\end{table}%

\subsection{Discussion}

The overall performance on each category is shown in Fig \ref{fig:detail_diff}. Although our method achieve promising results on the overall dataset, there are still some difficult actions could not be well recognized. A confusion matrix is shown in Fig \ref{fig:confuse}, where the selected difficult actions are determined according to the insufficient accuracies., e.g. less than $50\%$. One group of similar actions should be notices, including \emph{reading, writing, play with phone/tablet and type on a keyboard}. Only hands are slightly shaken in aforementioned actions, which makes them difficult to be distinguished. This issue is mainly caused by the lack of fine grained representation of two hands. Additionally, the noise contains in such area in the dataset. We believe that the performance could be improved by introducing body-object interaction information and noise refinement techniques in the future.

\section{Conclusions}\label{sec:conclusion}

To sum up, we presented a self-attention based skeleton-anchor proposal module in this paper. Instead of existing methods that conduct adjacency matrix focusing on pair-wise bone connection, the proposed SAP employ a high order triplet representation that encoding angle information, which is able to extract complementary feature to current ST-GCN based method.  In addition, the SAP module is able to automatically determine the anchor location via learning processing, where the anchor position could be located around body part. The ablation study has shown the advantage of proposed method compare with the manually specified strategy. By coupling with existing ST-GCN module, e.g. MS-G3D, our method achieves new state-of-the-art results on benchmark dataset. Besides, the proposed SAP module can be easily adopted to other curttent ST-GCN based methods to provide complementary information to current joint and bone representation.
{\small
\bibliographystyle{ieee_fullname}
\bibliography{angle_sequence}
}

\end{document}